\def\year{2020}\relax
\documentclass[letterpaper]{article}
\usepackage{aaai}
\usepackage{times}
\usepackage{helvet}
\usepackage{courier}
\usepackage{float}
\usepackage{url}
\usepackage{graphicx}
\usepackage{subcaption}
\usepackage{mwe}
\usepackage{algorithm}
\usepackage[noend]{algpseudocode}
\usepackage{multirow}
\usepackage{multicol}

\begin{document}
%
\title{Coarse2Fine: A Two-stage Training Method for Fine-grained Visual Classification}
\author{Amir Erfan Eshratifar,\textsuperscript{1}
         David Eigen,\textsuperscript{2}
         Michael Gormish,\textsuperscript{2}
         Massoud Pedram,\textsuperscript{1}\\
\textsuperscript{1}{Department of Electrical and Computer Engineering at University of Southern California, Los Angeles, CA 90089, USA}\\
\textsuperscript{2}{Clarifai, San Francisco, CA 94105, USA}\\
eshratif@usc.edu,
deigen@clarifai.com, 
michael.gormish@clarifai.com,
pedram@usc.edu}
\maketitle
\begin{abstract}
\begin{quote}
Small inter-class and large intra-class variations are the main challenges in fine-grained visual classification. Objects from different classes share visually similar structures and objects in the same class can have different poses and viewpoints. Therefore, the proper extraction of discriminative local features (e.g. bird's beak or car's headlight) is crucial. Most of the recent successes on this problem are based upon the attention models which can localize and attend the local discriminative objects parts. In this work, we propose a training method for visual attention networks, Coarse2Fine, which creates a differentiable path from the input space to the attended feature maps. Coarse2Fine learns an inverse mapping function from the attended feature maps to the informative regions in the raw image, which will guide the attention maps to better attend the fine-grained features. We show Coarse2Fine and orthogonal initialization of the attention weights can surpass the state-of-the-art accuracies on common fine-grained classification tasks.
\end{quote}
\end{abstract}

\section{Introduction}
\begin{figure}[t!]
\centering
\includegraphics{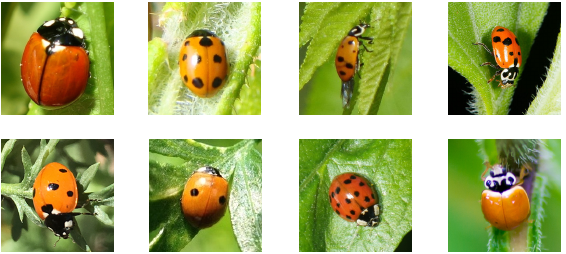}
\caption{Eight different lady bug species in iNaturalist dataset. There are very subtle differences between them like the number of spots.} \label{bugs}
\end{figure}

\noindent Fine-Grained Visual Classification (FGVC) aims to on differentiate hard-to-distinguish object classes. Common datasets on this task include different kinds of birds, cars, aircrafts, dogs. In FGVC, the differences between classes are usually very subtle but always visually measurable by humans. In addition to the large intra-class variations due to the variations of pose, lighting, and viewpoint, small inter-class variations is another challenge in FGVC. An example of the small inter-class variations is shown in Figure \ref{bugs}, in which the difference of these classes is only the number of spots and subtle appearance differences. Furthermore, because of the specialized domain knowledge requirements to label fine-grained classes in such datasets, there is often a lack of enough labeled data. As a result, because of these three challenges, it is usually hard to obtain accurate classification results only by the typical coarse-grained Convolutional Neural Networks (CNN).

As the fine-grained categories can be very similar to each other, extracting discriminative features from the object's parts is a key step. The current state-of-the-art in object's part feature learning can be divided into two group:  1. Methods which require part annotated datasets, 2. Methods which only need and image-level category label. In the first group, all the discriminative object parts are labeled with a location such as bounding boxes or landmarks \cite{ma-cnn}. As one can expect, location annotation of object parts is a cumbersome and expensive human labeling task and is also prone to human annotation errors. As a result, a body of research studies are focused on the design of fine-grained models with only image-level labeled data \cite{st-cnn,BilinearCNN,ra-cnn,partselection,ma-cnn}.

The core idea behind the image-level annotation-based methods is that only the image label itself supervises the model to predict the location of the object parts. Then, the local features are extracted from the predicted object's parts regions. Prediction of the object's parts locations with only image-level label requires an attention mechanism. The process of learning an attention mechanism is performed only on either feature space or input space. This paper studies the effect of making a connection between the attention mechanism of the feature and input spaces. Besides, another shortcoming of the current state-of-the-art fine-grained methods is that they are often limited in the number of object parts that they can predict (usually between 1-8 parts). As the object parts can be partially occluded, learning more than one part feature for a single object part can be potentially helpful because the model learns to extract appropriate features from the partially occluded parts. As we will show the proposed architecture can handle more part features because of the lower computational costs. The next issue is with the appropriate choice of the loss function. As the objective of the softmax cross-entropy loss is extracting the \textit{most} discriminative features, it can suffer from neglecting the less discriminative features which are crucial to learning for classifying the similar classes. 

Bilinear Attention Pooling (BAP) is a recently introduced attention mechanism for learning a set of attention maps for the most sensitive features \cite{WSBAN}. It is flexible for choosing a large number of discriminative object parts and been proven to improve the accuracy and part localization and is end-to-end trainable. BAP learns the attention maps for the feature maps without receiving any feedback from the corresponding attended region in the raw image. The improvements that can be achieved by matching the attended feature maps and their corresponding attended region in the raw image is studied by extensive experiments in this paper. 

\begin{figure}[H]
\centering
\includegraphics[]{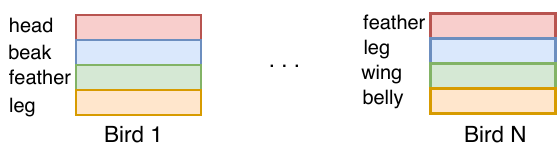}
\caption{The structure of the feature matrix using bilinear attention pooling network. Each row in the feature matrix will correspond to a discriminative object part. Each class has its discriminative parts and features which may differ from other classes. The attention mechanism itself learns these features with only using image-level category labels.} \label{feature_mat}
\end{figure}

As we will explain later, a loss function (Center Loss \cite{centerloss}) is used to penalize the discrepancy in feature values of the same object part. For example, learning a single feature vector for the beak of a certain bird species will help to learn more discriminative feature; while we do not have such ability by only using softmax cross-entropy loss. The feature matrix of bilinear attention models is presented in Figure \ref{feature_mat}. Each row of the class feature matrix is trained to extract a discriminant object part which might be different from other classes. This gives us the flexibility to train fine-grained models on different kind of species where the object parts can be different (e.g. birds, plants, reptiles, etc.). 

Also, the attention maps for a given class is desired to be as orthogonal as possible to each other to extract different object parts with fewer overlaps, which is less studied in the literature. As we will show, orthogonal initialization of the attention weights improves the accuracy by learning more discriminant object parts. 

The main contributions of this paper are as the following:
\begin{enumerate}
\item We propose a new deep model architecture for fine-grained feature extraction. The model is comprised of a cascade of two networks: coarse-grained and fine-grained networks. The attended features from the coarse-grained network are used to highlight the informative regions in the raw image which is then passed to the fine-grained network. The classification error from the fine-grained network on the highlighted image will guide the coarse-grained network to extract better features and attentions. The novelty of the architecture is the fact that it creates a feedback path from the raw input (highlighted image) to the attended feature maps. As the spatial location of activated neurons in the feature maps are highly correlated with the spatial location of the object's parts in the raw input domain, we expect that creating this feedback from the feature domain to input domain can help the model to localize and extract more accurate local discriminative features.
\item A learnable mapping function using a deconvolutional network from the attention maps to the raw pixel domain improving the state-of-the-art weakly supervised object localization. The prior art \cite{WSDAN} has used a linear up-sampler on the attention maps to localize the object in the image, which can be improved if replaced by a learnable up-sampler. The role of the up-sampler is to create a differentiable path between the attended features and their corresponding attended region in the raw image.
\item We experimentally show the effectiveness of orthogonal initialization of the attention module weights on the accuracy of the model. We expect each of attention map to be orthogonal to the other ones in the best scenario as extracting the most discriminant object parts is desired. For example, one may attend on the bird's beak and another one attends the bird's feather.
\end{enumerate}

\section{Related Work}
Convolutional Neural Networks (CNN) are designed to solve the large-scale image classification problem. As there is no specific module for recognizing subtle differences in the state-of-the-art CNNs such as ResNet \cite{resnet}, Inception \cite{inceptionv3}, and \cite{inceptionresnetv2}, they are all under-performing the specially crafted models for FGVC, which is shown in the Experiments and Result section. In this section, we study the related works about FGVC, bilinear pooling, object localization, and visual attention which are the core components of the proposed model. 

\textbf{Fine-grained visual classification} There is a body of research using the annotated part locations and attributes to supervise the model to be able to extract and focus on the local features. R-CNN and its extensions \cite{rcnn1,rcnn2} detect and localize the object parts and make predictions based on the pose-normalized features of the parts. \cite{deeplac} proposed a framework, Deep-LAC in which the alignment and classification errors are propagated back to the localization layers using their proposed valve linkage function (VLF). As the location labeling is a very expensive human workload, methods which only require an image-level category are drawing research attentions. \cite{st-cnn} proposed Spatial-Transformer CNNs which learns a proper geometric transformation by locating a few object's parts to align the image before the classification. \cite{ra-cnn} proposed a recurrent based CNN in which the model predicts the location of one attended region and extracts its corresponding feature, and then moves to the next part in the next time step. The recurrent nature of this model makes it computationally expensive as we need to run the inference $N$ times, where $N$ is the number of attention maps. To alleviate this issue, \cite{ma-cnn} proposed MultiAttention CNN which locates several attended regions instead of only one. However, the number of attention maps (2-4) in their model is still limited due to the computational costs. They also introduced a channel grouping loss for compact and diverse (orthogonal) part learning.

\textbf{Weakly supervised object localization} An object localization task in which we are not using any bounding boxes as labels during training is called Weakly Supervised Object Localization. Generally, weakly supervised learning referred to any learning process in which there are no direct labels for the targets \cite{10.1093/nsr/nwx106}. Indirect labels are used as weak supervisors to train the models. Localizing objects by using only their image-level label is a very challenging task. Prior arts \cite{discrim_localization} are based on generating localization maps using the activations in Global Average Pooling (GAP). As the location of the activated neurons in the output of GAP is highly correlated with the informative region in the image, they use an up-sampled copy of the GAP activations to localize the object. But, as many objects can be fully recognized with only a portion of their body, these models often fail to localize the whole object, and only localize a few object parts. To locate the whole object, \cite{ACoL} proposed Adversarial Complementary Learning (ACoL) approach to discover entire objects by training two adversary complementary classifiers, which can locate different object’s parts and discover the complementary regions that belong to the same object.

\textbf{Bilinear models} Bilinear pooling (BP) introduced by \cite{BilinearCNN} aggregates the spatial-wise outer-product of two features tensors from two CNNs by global pooling (average or max), as represented in Equation \ref{eq:bilinear_pooling}. Bilinear Attention Pooling (BAP) \cite{WSBAN} uses a similar idea to BP but the two feature tensors are coming from the same backbone model instead of two separate CNNs. In our work, we use BAP in which one stream is the feature map and the other one is the attention map which is supervised to learn the object's part distribution from the feature maps.

\textbf{Visual attention models} Visual attention learning proposed by \cite{wang17,wang18} learn an attention module for CNNs which assigns weights to different spatial locations in the raw image. It puts the attention module after each convolutional layer, which may be limiting the performance. Our proposed model is also a visual attention model which learns a set of attention maps which are supposed to attend both the feature maps and input image.

\section{Approach}
In this section, we elaborate on the proposed model. In summary, the proposed model architecture is comprised of the cascade of two feature networks which are equipped with attention learning mechanism by Bilinear Attention Pooling. The two cascaded networks are cascaded using a deconvolutional network. The attention maps are learned by adding a new loss function to the typical softmax cross entropy. We first elaborate on each component in the model individually then we'll explain the whole model. 

\subsection{Bilinear Attention Pooling}\label{BAP}
\begin{figure}[H]
\centering
\includegraphics[width=\columnwidth]{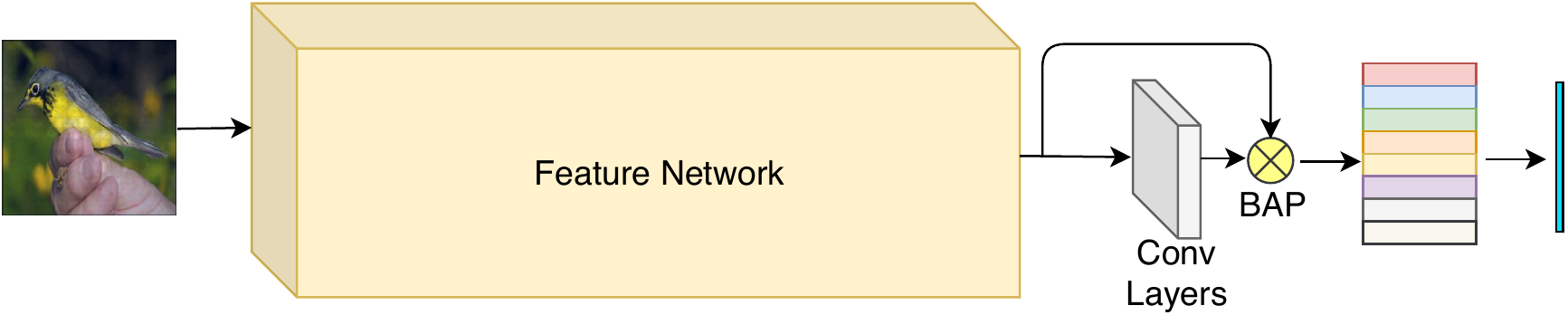}
\caption{Overview of the bilinear attention network's architecture. The difference of this architecture and a typical CNN is the extra BAP block, which is supposed to learn the attention maps for features. } \label{fig:wsdan}
\end{figure}

The idea of bilinear pooling first proposed by \cite{BilinearCNN}, incorporates two CNN models in parallel. The outer product of the two output feature tensors from these two models are computed and then for each location in the feature tensors, a pooling function (e.g. maximum, average, etc.) is applied across the channels. The pair-wise interaction between features is explicitly captured in this model architecture. For instance, the presence of a specific car's headlight and windshield and their interaction can be helpful for more accurate classification. But the issue with this approach is the expensive computational cost due to the presence of two CNN models. To alleviate the costs of two parallel networks, one can create a parallel branch from a selected layer in a single network. The closer it is to the classification layer, the less computational overhead it causes. As depicted in Figure \ref{fig:wsdan}, in our studied model, the first stream is the output features of the backbone feature network, feature maps, and the second one is obtained after one or several convolutional layers, which we call attention maps. The attention maps are supervised in a way to learn the object's part distributions, which we will explain later.

Given two feature tensors $f_1$ and $f_2$, the Bilinear Pooling (BP) operation on them can be written as:

\begin{equation}
BP = pool(\{ f_1^i {f_2^i}^T \})_{i=1}^{i=H \times W}
\label{eq:bilinear_pooling}
\end{equation}

\begin{figure}[H]
\centering
\includegraphics{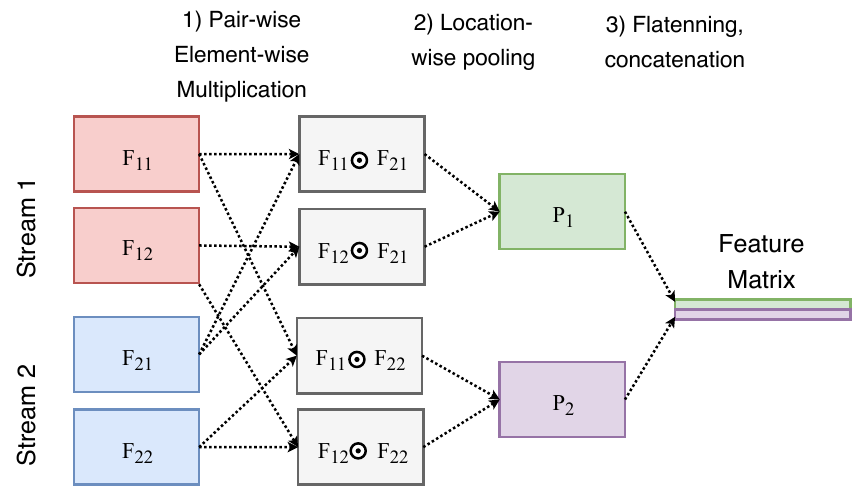}
\caption{Bilinear feature pooling for two streams of features with two channels. Each feature matrix in a stream is element-wise multiplied to all the feature matrices of the other stream. Then, a pooling function (e.g. average, max) is applied at each location and the resulting feature matrices are flattened and concatenated. Each row in the final feature matrix is supposed to extract a different object part (e.g bird's head, car's headlight)
} \label{bilinear_pooling}
\end{figure}

We assume both $f_1$ and $f_2$ have the same spatial size of $H \times W$ and their channel size will be $N$ and $M$, respectively. The pooling function is applied on each spatial location ($i=1:H \times W$) across all $N \times M$ channels. In our case, assume $f_1$ as the feature maps with $N$ channels and $f_2$ as the attention maps with $M$ channels. The difference between the Bilinear Attention Pooling (BAP) and BP is in the pooling function. As we aim to train a set of attention maps so that each of them is attending a different object part, pooling is applied $M$ times over the $N$ channels, separately. This means that with BAP, we obtain a tensor of size $M \times H \times W$ instead of $1 \times H \times W$, which is the output size of BP. As a result, with each attention map ($\#M$), we obtain a different feature vector which is supposed to represent a specific object part. This operation is demonstrated in Figure \ref{bilinear_pooling} for $N=M=2$. After the pooling step, we flatten the $M \times H \times W$ tensors into $M \times H*W$ matrix, where each row represents different object's part.

\begin{figure*}[t]
\centering
\includegraphics[width=\linewidth]{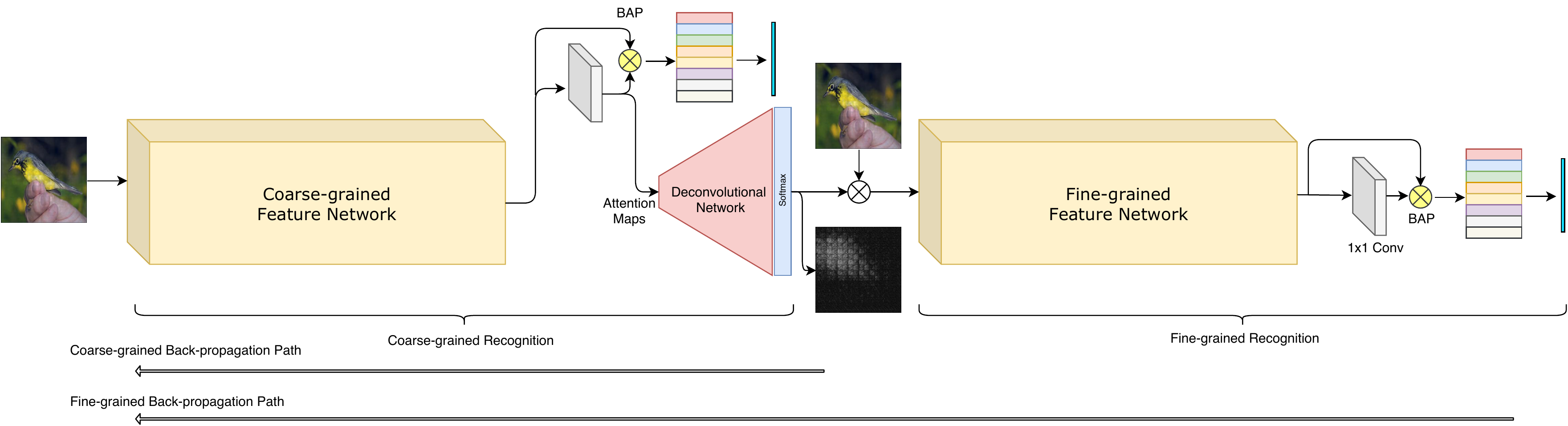}
\caption{Overview of the proposed training method. The coarse-grained network generates a set of attention maps on the feature space and also the coarse-grained label predictions. The attention maps are up-sampled using a deconvolutional network and are used as a set of masks to highlight the informative regions in the image. The highlighted image is passed through another network which is the fine-grained specialist and fine-grained predictions are made. The back-propagation from the fine-grained classification error to the generated attention maps supervises them to attend the features which will reduce the fine-grained classification error. } \label{proposed_model}
\label{training_results}
\end{figure*}

\subsection{Attention Learning}\label{attention_learning}
As the objective of the Softmax cross-entropy is to minimize only the classification risk, there is not any force on the attention maps to attend only the same object parts. To alleviate this issue, Center Loss \cite{centerloss} is used upon the feature matrix. In this way, for each row in the feature matrix, which is a specific object's part feature (e.g. beak), a center is learned and features are enforced to get closer to that center with a regularization like L2. Therefore, we will have $C$ center matrices of size $M \times H*W$ for all classes. If we represent the feature matrix of $i$th class with $f_i$ and center matrix of $i$th class with $c_i$, the loss function for attention learning will be:

\begin{equation}
L_{attention} = \frac{1}{C}\sum_{i=1}^{C} \left\Vert f_i - c_i \right\Vert^2
\end{equation}

The centers are updated using moving average during the training:

\begin{equation}
c_i = (1 - \beta) c_i + \beta f_i
\end{equation}

where $\beta$ acts as the momentum value of the updates. We set the starting point of the center matrices to zero.
\subsection{Coarse2Fine Training}\label{Coarse2Fine}
As depicted in Figure \ref{proposed_model}, the proposed model is comprised of two feature networks, BAP, and a deconvolutional network. The first feature network takes the input image and outputs the attention maps and a coarse-grained prediction. During the training, we take one of the attention maps randomly. The selected attention map is passed into an up-sampler to create a mask over the attended input image regions. Then, the selected up-sampled attention map and the input image are element-wise multiplied into each other to create a new image in which a sensitive region is highlighted. Then, this image is passed through another network which is always receiving the highlighted images from the coarse-grained step as the input and learns a classifier based on those highlighted regions. As a result, we expect the second network to be an expert in fine-grained classification. In a nutshell, the whole network has two classification heads, one for coarse-grained prediction and one for fine-grained prediction and is end-to-end trainable. The feedback from the fine-grained classification part learns a mapping from the attended feature maps (attention maps) to their associated attended input image regions. As the spatial locations of the activated neurons in the attention maps are highly correlated with the informative regions in the raw input image, one could also use a typical non-trainable up-sampler (e.g. bilinear) but should expect some accuracy degradation \cite{WSDAN}.

As we explained in the previous the loss function is the combination of two softmax cross entropy loss functions for coarse-grained and fine-grained classification and center loss for attention learning:
\begin{equation}
L = L_{coarse-grained} + L_{fine-grained} + \lambda L_{attention}
\end{equation}
where $\lambda$ is the regularization factor for the attention loss. 

\textbf{Orthogonal initialization of the attention weights} As each of the attention maps are supposed to extract an object part feature, we expect them to be as orthogonal as possible to each other. As the attention weights is a 1x1 convolution layer, it can be represented as a single 2d matrix, $X$. If we assume its SVD decomposition as $X=U \Sigma V^T$, we take the matrix $V$ as the initial weights for the attention weights. We also added L1 and L2 regularization on the pairwise dot product of the attention maps. It didn't perform well but will be studied more in detail as future work.

The whole training process is shown in Algorithm \ref{coarse2fine_algorithm}. 

\textbf{Inference} The random selection of the attention maps during the training is replaced by taking the mean of all attention maps during the inference. As a result, the model will see all the attended regions instead of one for doing the inference. Also, as we have two prediction heads there are three options for making the predictions: 1- Only coarse-grained, 2- Only fine-grained, 3- Average of coarse and fine-grained. If the performance matters, one can only use the coarse-grained part which itself performs better than a typical CNN. Our experiments showed that the average of both coarse and fine-grained predictions results in the best accuracy.

\begin{algorithm}
\caption{Coarse2Fine training algorithm}\label{coarse2fine_algorithm}
\begin{algorithmic}[1]  
	\State {Initialize model parameters, $\theta$}
	\State $U$, $\Sigma$, $V^T$ = SVD($AttentionWeights$)
	\State $AttentionWeights = V$ 
     \For{\texttt{iteration = 1,2,...}}
        \State Sample a mini-batch of images $B_i$
        \State CoarsePreds, Attns, CoarseFeats = CoarseNet($B_i$)
        \State UpAttns = Deconv(Attns)
        \State Select a random attention map ($A_k$) from UpAttns
        \State FinePreds, FineFeats = FineNet($A_k * B_i$)
        \State $L_1$ = $L_{cross-entropy}$(CoarsePreds)
        \State $L_2$ = $L_{cross-entropy}$(FinePreds)
        \State $L_3$ = $L_{center-loss}$(CoarseFeats+FineFeats)
        \State $L$ = $L_1$ + $L_2$ + $L_3$
        \State $\theta$ = $\theta$ - $\alpha$ $\nabla L$
      \EndFor
	\State \Return $\theta$
\end{algorithmic}
\end{algorithm}

\section{Experiments and Results}\label{experiments}
In this section, we present the experimental setups and results to demonstrate the effectiveness of the Coarse2Fine training method. We compare Coarse2Fine performance on five publicly available datasets to the state-of-the-art fine-grained methods.

\subsection{Datasets}\label{datasets}
In the FGVC literature, there are four common datasets including CUB-200-2011 \cite{CUB_200_2011}, FGVC-Aircraft \cite{AirCraft}, Stanford Cars \cite{stanford_cars}, Stanford Dogs \cite{stanford_dogs}. One less studied but challenging dataset in FGVC is iNaturalist \cite{inaturalist}. It includes many similar looking species which are captured from various locations in the world. The training, test, and class size of all these datasets are shown in Table \ref{tab:datasets}.

\subsection{Setup}\label{experimental_setup}
We use InceptionV3 \cite{inceptionv3} as the backbone feature network. To reduce the costs of having two separate networks, we shared\ the feature network for both coarse-grained and fine-grained parts. The output of "Mix6e" layer in InceptionV3 is used as the feature maps. Attention maps are obtained by a $1\times 1$ convolution from the feature maps. The number of attention maps is set to 8, $\lambda$ is set to 1.0 and we train the models using Stochastic Gradient Descent (SGD) with the momentum of 0.9, epoch number of 200, weight decay of 0.00001, and mini-batch size of 64 on four GTX 1080 Ti GPU.

\textbf{Pre-training of the deconvolutional network} As our up-sampler network is sitting in between of two already pre-trained models, starting from random initialization point could adversely affect the coarse-grained model by feeding noisy gradients during the early phase of training. As a result, we trained the up-sampler separately on our synthetic dataset. We created a dataset of pair of 60000 input and its up-sampled version using bilinear, cubic, and nearest interpolators. We trained the up-sampler network using L2 loss for 100 epochs.

\subsection{Fine-grained Visual Classification Results}\label{fgvc_results}
The results of the FGVC using our proposed method is shown for 5 different FGVC datasets in Table \ref{tab:birds_results}, Table \ref{tab:aircraft_results}, Table \ref{tab:cars_results}, Table \ref{tab:dogs_results}, Table \ref{tab:inat_results}. Coarse2Fine achieves the state-of-art
performance on all these fine-grained datasets. The interesting point is that the final model contains only 6144 extra parameters, which is almost nothing compared to 24M total parameters in InceptionV3.

\begin{table}[H]
\caption{Fine-grained datasets specifications}
\centering
\label{tab:datasets}
\begin{tabular}{|c|c|c|c|c|}
\hline
Dataset & Objects & Categories & Training & Test \\ \hline
CUB-200-2011 & Bird & 200 & 5974 & 5794                \\ \hline
FGVC-Aircraft & Aircraft & 100 & 6667 & 3333                \\ \hline
Stanford Cars & Car & 196 & 8144 & 8041                \\ \hline
Stanford Dogs & Dog & 120 & 12000 & 8580                \\ \hline
iNaturalist 2017 & Species & 5089 & 12000 & 8580                \\ \hline
\end{tabular}
\end{table}

As we will explain in the discussion section, we also measure the effectiveness of our fine-grained model on face attributes. For this purpose, we use CelebA dataset \cite{celeba} which includes 40 binary attributes of the human face.

\begin{table}[H]
\caption{Classification results on CUB-200-2011 dataset.}
  \centering
\label{tab:birds_results}
\begin{tabular}{c|c}
\hline
Model             & Top 1 Accuracy (\%) \\ \hline
ResNet-101 \cite{resnet}       & 83.5                \\ 
InceptionV3 \cite{inceptionv3}      & 83.7                \\ 
IncResNetV2 \cite{inceptionresnetv2} & 84.0                \\ \hline
PA-CNN \cite{pa-cnn}           & 82.8                \\ 
B-CNN \cite{BilinearCNN}            & 84.1                \\ 
ST-CNN \cite{st-cnn}           & 84.1                \\ 
RA-CNN \cite{ra-cnn}           & 85.4                \\ 
GP-256 \cite{gp-256}          & 85.8                \\  
MA-CNN \cite{ma-cnn}           & 86.5                \\  
MAMC \cite{mamc}             & 86.5                \\ 
PC \cite{pc}               & 86.9                \\ 
DFL-CNN \cite{dfl-cnn}     & 87.4                \\
NTS-NET \cite{nts-net}     & 87.5                \\\
MPN-COV \cite{mpn-cov}           & 88.7                \\ 
WS-DAN \cite{WSDAN}           & 89.4                \\ \hline
\textbf{Coarse2Fine}       & \textbf{89.5}               
\end{tabular}
\end{table}

\begin{table}[H]
\caption{Classification results on FGVC-Aircraft dataset.}
  \centering
\label{tab:aircraft_results}
\begin{tabular}{c|c}
\hline
Model             & Top 1 Accuracy (\%) \\ \hline
ResNet-101 \cite{resnet}       & 87.2                \\ 
InceptionV3 \cite{inceptionv3}      & 87.4                \\ 
IncResNetV2 \cite{inceptionresnetv2} & 88.1                \\ \hline

B-CNN \cite{BilinearCNN}            & 84.1                \\ 
RA-CNN \cite{ra-cnn}           & 88.4                \\
PC \cite{pc}               & 89.2                \\ 
GP-256 \cite{gp-256}          & 89.8                \\  
MA-CNN \cite{ma-cnn}           & 89.9                \\  
MPN-COV \cite{mpn-cov}           & 91.4                \\ 
NTS-NET \cite{nts-net}     & 91.4                \\
DFL-CNN \cite{dfl-cnn}     & 92.0               \\

WS-DAN \cite{WSDAN}           & 93.0                \\ \hline
\textbf{Coarse2Fine}       & \textbf{93.4}               
\end{tabular}
\end{table}

\begin{table}[H]
\caption{Classification results on Stanford Cars dataset.}
  \centering
\label{tab:cars_results}
\begin{tabular}{c|c}
\hline
Model             & Top 1 Accuracy (\%) \\ \hline
ResNet-101 \cite{resnet}       & 91.2                \\ 
InceptionV3 \cite{inceptionv3}      & 90.8                \\ 
IncResNetV2 \cite{inceptionresnetv2} & 91.5                \\ \hline

RA-CNN \cite{ra-cnn}           & 92.5                \\ 
MA-CNN \cite{ma-cnn}           & 92.8                \\  
GP-256 \cite{gp-256}          & 92.8                \\ 
PC \cite{pc}               & 92.9                \\ 
MAMC \cite{mamc}             & 93.0                \\ 
MPN-COV \cite{mpn-cov}           & 93.3                \\ 
DFL-CNN \cite{dfl-cnn}     & 93.8                \\
NTS-NET \cite{nts-net}     & 93.9                \\

WS-DAN \cite{WSDAN}           & 94.5                \\ \hline
\textbf{Coarse2Fine}       & \textbf{94.7}               
\end{tabular}
\end{table}

\begin{table}[H]
\caption{Classification results on Stanford Dogs dataset.}
  \centering
\label{tab:dogs_results}
\begin{tabular}{c|c}
\hline
Model             & Top 1 Accuracy (\%) \\ \hline
ResNet-101 \cite{resnet}       & 85.8                \\ 
InceptionV3 \cite{inceptionv3}      & 88.9                \\ 
IncResNetV2 \cite{inceptionresnetv2} & 90.0                \\ \hline

NAC \cite{NAC}               & 68.6                \\ 
PC \cite{pc}               & 83.8                \\ 
FCAN \cite{pc}               & 84.2                \\ 
MAMC \cite{mamc}             & 85.2                \\ 
RA-CNN \cite{ra-cnn}           & 87.3               \\ 

WS-DAN \cite{WSDAN}           & 92.2               \\ \hline
\textbf{Coarse2Fine}       & \textbf{93.0}               
\end{tabular}
\end{table}

\begin{table}[H]
\caption{Classification results on iNaturalist 2017 dataset.}
  \centering
\label{tab:inat_results}
\begin{tabular}{c|c}
\hline
Model             & Top 1 Accuracy (\%) \\ \hline
ResNet-101 \cite{resnet}       & 58.4                \\ 
InceptionV3 \cite{inceptionv3}      & 64.2                \\ 
IncResNetV2 \cite{inceptionresnetv2} & 67.3                \\ \hline
WS-DAN \cite{WSDAN}           & 68.9                \\ \hline
\textbf{Coarse2Fine}       & \textbf{70.5}               
\end{tabular}
\end{table}

\subsection{Object Localization Results}\label{localization_results}

\begin{figure}[H]
\centering
\includegraphics[width=\columnwidth]{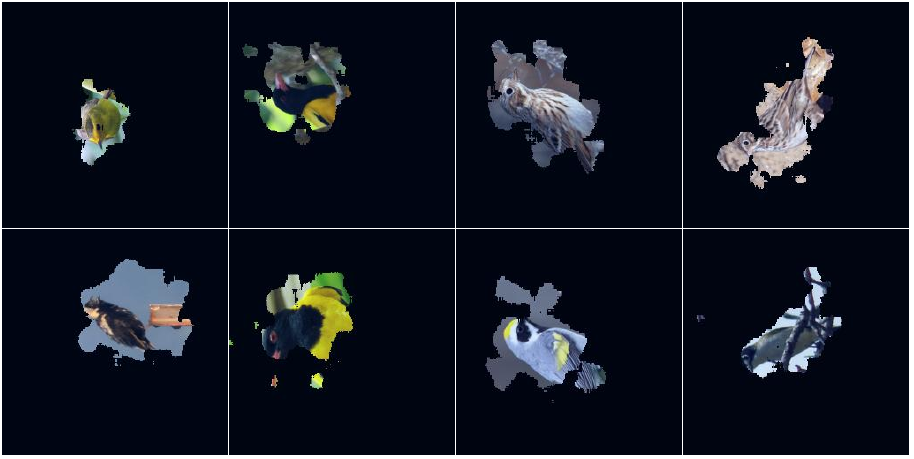}
\caption{Visualization of the object localization using the attention maps. We passed the attention maps through the deconvolutional up-sampler, then an averaging convolution followed by a threshold function is performed to mask the attended region. In our object localization, we do not use any bounding boxes during the training. }
\label{birds_attention}
\end{figure}

The weakly supervised object localization is a recently attractive area of research. The term "weakly" supervised means we do not use any bounding box but just the image-level labels. As most of the objects in Stanford Cars and FGVC-Aircrafts occupy almost the whole image, we evaluate on CUB-200-2011 and Stanford Dogs where the objects are quite smaller than the whole image size. To have a fair comparison, we use the localization metric introduced in \cite{ACoL}. In this metric, the localization is correct if the Intersection over Union (IoU) of the true bounding box and predicted one is greater than 50\% and the predicted label is correct. We selected eight random bird images from the test set and visualized the output of the attention maps which is depicted in Figure \ref{birds_attention}.

\begin{table}[H]
\caption{Object localization errors on CUB-200-2011 and Stanford Dogs datasets.}
  \centering
\label{tab:birds_local_results}
\begin{tabular}{c|c|c}
\hline
Model             & CUB-200-2011 (\%) & Stanford Dogs\\ \hline
GoogLeNet       & 59.0  & 30.7              \\ 
VGGnet-ACoL      & 54.1   & -              \\ 
ResNet-101 & 42.1   & 29.6              \\ 
InceptionV3      & 40.8   & 28.8              \\
WS-DAN          & 18.3  & 19.2               \\ \hline
\textbf{Coarse2Fine}       & \textbf{17.2}  & \textbf{17.9}              
\end{tabular}
\end{table}

\subsection{Ablation Study: Orthogonal Initialization}\label{ablation_studies}
To demonstrate the effect of orthogonal initialization of attention weights, we provide the accuracy results of Coarse2Fine with and without the orthogonal initialization in Table \ref{tab:ablation_ortho}.

\begin{table}[H]
\caption{Ablation study of the orthogonal initialization of attention weights using Coarse2Fine model.}
  \centering
\label{tab:ablation_ortho}
\begin{tabular}{|c|c|c|c|}
\hline
Dataset             &  With ortho init & Without ortho init\\ \hline
CUB-200-2011        & 89.5 & 89.3             \\ \hline
FGVC-Aircrafts     &  93.4  & 93.1             \\  \hline
Stanford Cars  &  94.7   & 94.5        \\ \hline
Stanford Dogs  & 93.0  & 92.6         \\ \hline
iNaturalist 2017  & 70.5 & 69.7  \\ \hline           
\end{tabular}
\end{table}

As the desired expectation is to obtain a set of attentions that each only attend a specific object part the orthogonality of them can be crucial. We also added L1 and L2 regularization methods on the pairwise dot products of the attention maps which reduced the accuracy but can be more investigated as a future work.

\section{Discussion} \label{discussion}
As the fine-grained classification techniques are designed to extract and discriminate subtle features between objects, their applications to face recognition is an interesting case study. As we could have possibly billions of humans faces to classify, the inter-class differences could be very small. In this section, we demonstrate the improvements can be achieved in the face domain using the proposed fine-grained architecture. We performed experiments on CelebA dataset which includes 40 binary attributes of the human's face (e.g. smiling, attractive, etc.). We selected 6 attributes which nearly split the dataset into the half positive and half negative. We trained a typical InceptionV3 with and without the attention layers. The number of attention maps is set to 8 which will lead to only $8*768=6144$ extra parameters compared to the total 24M parameters of InceptionV3. All the other configurations like hyperparameters, initialization, epochs, etc. are the same for both networks. As we have shown in Figure \ref{tab:celeba_results}, the accuracy of the model with attention is improved by 2.1\% compared to the model without attention.

\begin{table}[H]
\caption{Accuracy Results on CelebA dataset}
  \centering
\label{tab:celeba_results}
\begin{tabular}{|c|c|c|c|}
\hline
Attribute   & IncV3 & IncV3 + Attention\\ \hline
Attractive  & 80.84 & 84.01 \\ \hline
High Cheekbone  & 83.50 & 85.23 \\ \hline
Male  & 91.18 & 92.88 \\ \hline
Mouth Slightly Open & 90.95 & 92.31 \\ \hline
Smiling & 92.96 & 95.67 \\ \hline
Wearing Lipstick & 86.51 & 88.85 \\ \hline
\end{tabular}
\end{table}

\section{Conclusion and Future Work}
In this paper, we proposed a training method for improving the local discriminative feature learning which is the key ingredient for fine-grained visual classification. The proposed method is based on attention learning which is supervised to attend both the feature maps and the raw input image by creating a feedback from the raw input space to the attention maps. Orthogonal initialization of the attention maps helps the attention maps to learn different object's parts and reduce the overlap between them. The proposed architecture and the orthogonal initialization achieve the state-of-the-art performance in fine-grained visual classification datasets.

As a future work, as the problem of large-scale face recognition can be considered as a fine-grained classification, the effectiveness of the fine-grained techniques can be further studied on face datasets. Creating datasets which include the most confusing human faces can further boost the progress in the fine-grained feature extraction.

\section{Acknowledgments}
This work has been done during the internship of Amir Erfan Eshratifar at Clarifai.

\bibliographystyle{aaai}
\bibliography{references.bib}

\end{document}